\documentclass[conference]{IEEEtran}
 
\IEEEoverridecommandlockouts 
\usepackage{cite}
\usepackage{amsmath,amssymb,amsfonts}
\usepackage{algorithmic}
\usepackage{graphicx}
\usepackage{textcomp}
\usepackage{xcolor}

\usepackage{tabularray}
\usepackage{array}
\usepackage{graphicx}
 
\usepackage{tabularray}
\usepackage{stfloats}
\DeclareUnicodeCharacter{FF08}{\lparen}
\DeclareUnicodeCharacter{FF09}{\rparen}
\DeclareUnicodeCharacter{FF1A}{:}
\newcommand{\TableNote}[1]{\textbf{Note:} #1}

\def\BibTeX{{\rm B\kern-.05em{\sc i\kern-.025em b}\kern-.08em
    T\kern-.1667em\lower.7ex\hbox{E}\kern-.125emX}}
\begin{document}

\title{Implicit  Sentiment Analysis Based on  Chain-of-Thought Prompting
}

\author{\IEEEauthorblockN{Zhihua Duan}
\IEEEauthorblockA{\textit{Intelligent Cloud Network Monitoring Department  } \\
\textit{China Telecom Shanghai Company}\\
Shanghai, China \\
* Corresponding author: duanzh.sh@chinatelecom.cn}
\and
\IEEEauthorblockN{Jialin Wang}
\IEEEauthorblockA{\textit{Department of Computer Science   } \\
\textit{Stanford University}\\
California, America \\
jialinwangspace@gmail.com} 
}

\maketitle

\begin{abstract}
Implicit Sentiment Analysis (ISA) is a crucial research area in natural language processing. Inspired by the idea of large language model Chain of Thought(CoT), this paper introduces a Sentiment Analysis of Thinking (SAoT) framework. The framework first analyzes the implicit aspects and opinions in the text using common sense and thinking chain capabilities. Then, it reflects on the process of implicit sentiment analysis, and finally, it deduces the polarity of sentiment. The model is evaluated on the SemEval 2014 dataset, consisting of 1120 restaurant reviews and 638 laptop reviews. The experimental results demonstrate that the utilization of the ERNIE-Bot-4+SAoT model yields a notable performance improvement. Specifically, on the restaurant dataset, the F1 score reaches 75.27, accompanied by an ISA score of 66.29. Similarly, on the computer dataset, the F1 score achieves 76.50, while the ISA score amounts to 73.46. Comparatively, the ERNIE-Bot-4+SAoT model surpasses the BERT$_{Asp}$+SCAPT baseline by an average margin of 47.99\%. 
\end{abstract}


\section{Introduction}
The sentiment analysis of text can be performed through natural language processing (NLP), including positive, negative, or neutral. It enables the understanding of customers language, expression, and underlying emotional meaning.There are many subtasks and topics in sentiment analysis.Implicit sentiment analysis, As part of sentiment analysis,aims to identify implicitly expressed sentiment viewpoints and polarities. 

Based on the presence or absence of explicit sentiment words, it can be categorized into Explicit Sentiment Analysis (ESA) and Implicit Sentiment Analysis (ISA). ESA is the mainstream research scenario, where explicit sentiment expressions exist in the text. Compared to ESA, ISA faces the following complex challenges:1. Lack of explicit sentiment words: ISA lacks explicit sentiment words or expressions, making it difficult to identify and extract potential sentiments from the text. Implicit sentiments are often conveyed through subtle cues, nuanced language, metaphors, or context, requiring a deeper understanding of the text.
2. Context and subjective expressions: Implicit sentiments are heavily influenced by the context in which they occur. Understanding the context and its impact on sentiment requires analyzing a broader range of language and situational factors surrounding the text. Subjective expressions further increase the complexity of analysis as they involve personal viewpoints, opinions, and individual experiences.3. Ambiguity and polysemy: Implicit sentiments can be ambiguous, with multiple possible interpretations. Words or phrases may have different meanings and can be associated with various sentiments depending on the context.  

Fortunately, in recent years, A Transformer-based approach to large language models\cite{attention-need},such as ChatGPT, LLaMA\cite{llama,llama2}, and GPT4 have made significant advancements, providing us with a promising solution to our problem. On one hand, LLMs through extensive pre-training, have acquired vast amounts of world knowledge and demonstrated outstanding capabilities in common-sense understanding, effortlessly tackling common-sense reasoning tasks. On the other hand, The latest insights in Chain of Thought (CoT) theory underscore the extensive capabilities of LLMs in analytical thinking.In CoT reasoning, by appropriately guiding the LLM, it can engage in chain-based reasoning\cite{prompt-CoT}. 
 
To address the challenges of implicit sentiment analysis,This investigation was stimulated by the CoT principle in large language models(LLMS) and introduces a new sentiment analysis framework called Sentiment Analysis of Thought (SAoT).

\section{Related Work}

 Sentiment analysis techniques vary in levels and methods, and they possess different levels of task complexity. Rule-based and lexicon-based methods require pre-defined sentiment lexicons and manual annotation of texts, incurring significant labor and time costs. Rule-based approaches tend to exhibit limited effectiveness when dealing with intricate or heterogeneous textual data, as they are restricted to recognizing only elementary sentiments or topics.Machine learning-based methods rely on manually designed sentiment features, which often yield suboptimal results. Using deep learning methods requires a lot of training data and a lot of computing power, and they have limitations in sentiment recognition in texts, particularly in handling complex emotions and topics\cite{survey-SA-2022}.

Prompt-based sentiment analysis methods: Recent research has highlighted the potential of prompt engineering to enhance language models' reasoning abilities.By generating prompt templates, language models can improve their performance on  reasoning and commonsense reasoning tasks \cite{prompt-CoT}.they have failed to critically reflect on the limitations of prompt-based sentiment analysis methods.

Integrating language models with knowledge bases or environments can produce better synergistic effects \cite{ReAct}. Modular systems combining neural networks and symbolic methods have also been studied \cite{mrkl}. prior works have not fully explored the integration of sentiment analysis systems.In terms of applications,The concept of prompt learning has gained significant attention in the field of natural language processing, as demonstrated in a recent survey \cite{survey-prompt}.
In the field of sentiment analysis, while sentiment analysis systems aim to determine the sentiment polarity of a given target based on explicit opinion expressions in the input text, implicit sentiment analysis (ISA) focuses on opinion clues that appear implicitly and ambiguously. The THOR CoT framework is a novel approach to implicit sentiment detection, utilizing multi-hop reasoning to gradually induce implicit aspects and opinions \cite{prompt-thor}.Prior studies have indicated that prompt tuning enhances language model performance \cite{prompt-thor, prompt-domain}. however,the works fail to critically reflect on the limitations of the prompt used.

Our research offers solutions to address these limitations. By designing a Sentiment Analysis of Thoughts (SAoT) framework and formulating effective prompt design principles and cues, we guide large-scale language models in analyzing implicit aspects and opinions, conducting critical reflections on implicit sentiment analysis, and inferring overall emotional polarity. By bridging these gaps, our findings contribute to the advancement of this field.

\section{SAoT Framework} 
As illustrated in Figure 1, the framework leverages common-sense knowledge and the ability to form chains of thought to conduct in-depth analysis of implicit aspects and viewpoints in the text. By reflecting on the process of implicit sentiment analysis, the framework infers the polarity of sentiments.
\begin{figure*}[htbp]
    \centering
    \includegraphics[width=0.89\textwidth]{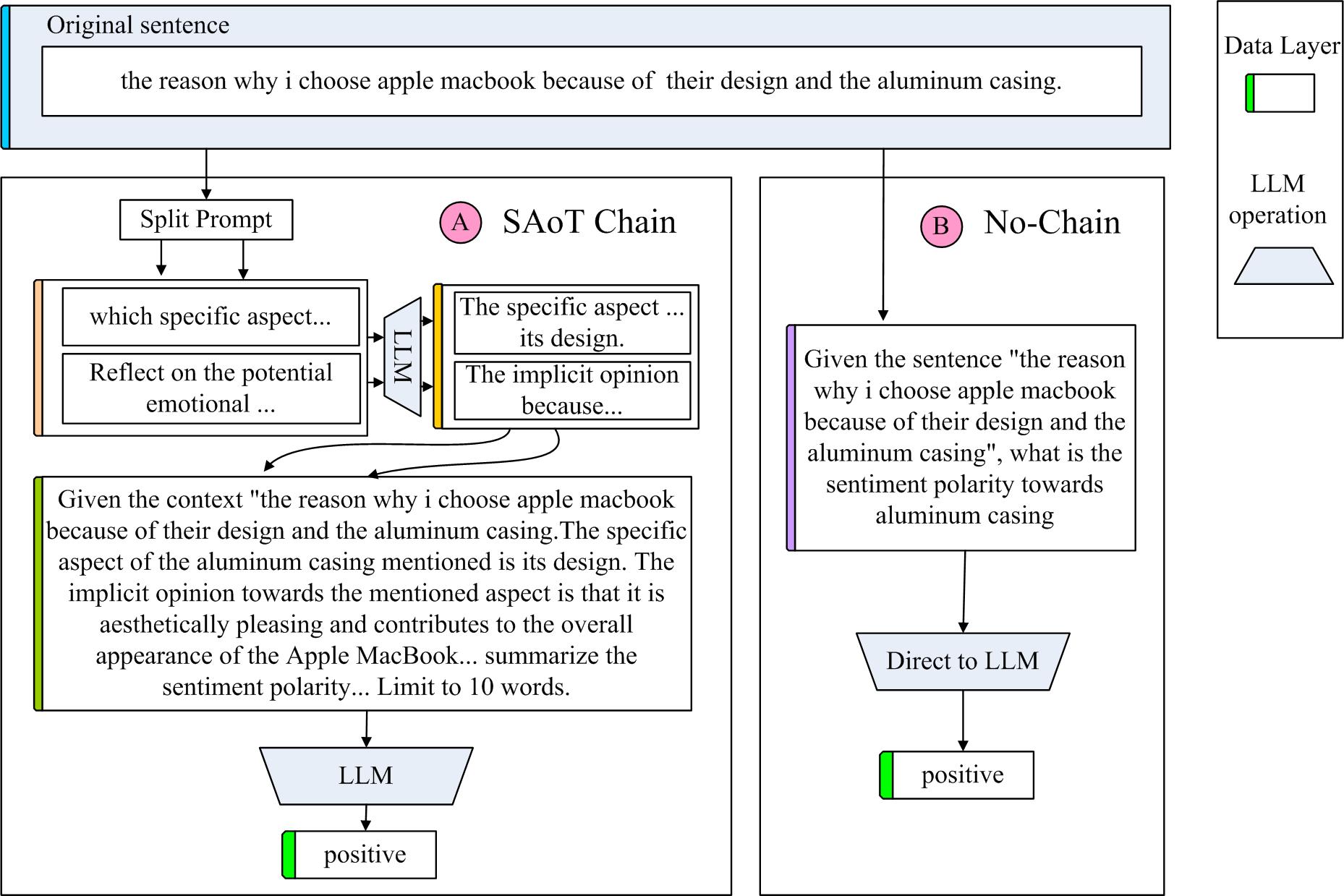}
    \caption{SAoT Framework Architecture Diagram}
    \end{figure*} 

In Figure 1, A: SAoT Chain is depicted, which utilizes split prompts. One prompt is used for conducting in-depth analysis of implicit aspects and viewpoints in the text, while the other prompt is used for reflecting on the process of implicit sentiment analysis. Their respective outputs are obtained by applying a language model. Subsequently, The outcomes are concatenated and forwarded together to the language model to attain the final result.B: No-Chain, on the other hand, utilizes direct input of prompts into a language model to acquire the ultimate outcome.

The key findings of this paper are as follows:We propose a sentiment analysis inference framework based on the CoT concept and design a set of effective prompt words to analyze implicit sentiment viewpoints in the text.
\begin{itemize}
\item Firstly, we utilize common-sense knowledge and the ability to form chains of thought to analyze implicit aspects and viewpoints in the text. Through analyzing the context  and relevant information, we can capture the implicit sentiment viewpoints that are not explicitly expressed, enriching the content of sentiment analysis.

\item Secondly, we reflect on the process of implicit sentiment analysis. Traditional methods often focus only on the surface-level sentiment information in the text, overlooking the underlying sentiment intentions. We believe that implicit sentiment analysis requires further thinking and reasoning to uncover the deeper meanings behind the sentiments. By introducing the process of inference, we can more accurately infer the sentiment polarity in the text, providing more comprehensive and accurate results for sentiment analysis.

\item Furthermore, we deduce the sentiment polarity results. By combining the semantic information of the text with the sentiment targets, we can better understand the meaning of the sentiment expressions and deduce the corresponding sentiment polarity.
\end{itemize}
In the experimental phase, we combine the SAoT framework with the ERNIE-Bot-4 model and conduct a detailed evaluation and comparison in a zero-shot setting. The experimental results demonstrate that the SAoT+ERNIE-Bot-4 framework achieves significant performance improvements in sentiment analysis tasks.

\section{Experiments}
\subsection{Dataset} This study aims to conduct experiments on zero-shot sentiment analysis, where we can predict sentiment even without fine-tuning or pre-training by using large-scale models and prompt words. 
Two widely used datasets, the Laptop and Restaurant datasets, were selected for our experimental analysis.These datasets are sourced from SemEval2014 and are among the most commonly used sentiment analysis datasets. For this research, we exclusively utilized the test data from these datasets, which include user review texts, targets, indications of implicit sentiment, and corresponding sentiment polarity labels (positive, negative, neutral). More detailed information about the datasets can be found in Table 1. By conducting experiments on these datasets, Our objective is to investigate the efficiency of the SAoT approach in the zero-shot sentiment analysis task and analyze its performance. 
 
\begin{table}[h]
\centering
\caption{Data distribution}
\begin{tblr}{
  width = \linewidth,
  colspec = {Q[190]Q[162]Q[140]Q[137]Q[96]Q[75]Q[123]},
  hlines,
  hline{1,4} = {-}{0.08em},
}
DataSet & Negative & Positive & Neutral & Total & ISA & ISA(\%)\\
Restaurant & 196 & 728 & 196 & 1120 & 267 & 23.83\\
Laptop & 128 & 341 & 169 & 638 & 175 & 27.42
\end{tblr}
\end{table}

We utilized a subset of the SemEval 2014 dataset, which consists of two subsets: Restaurant and Laptop. The Restaurant dataset comprises a total of 1120 instances, with 267 instances (23.83\%) labeled as having implicit sentiment. Within this dataset, there are 196 instances labeled as negative, 728 instances labeled as positive, and 196 instances labeled as neutral.The Laptop dataset contains 638 instances, with 175 instances (27.42\%) having implicit sentiment. Among these instances, there are 128 instances labeled as negative, 341 instances labeled as positive, and 169 instances labeled as neutral.

\subsection{Setup} In this research, we performed experiments on two publicly available datasets, specifically the Laptop and Restaurant datasets from SemEval14. These datasets consist of explicit and implicit sentiment records. We utilized Flan-T5 as the underlying network for the Large Language Model (LLM) and tested it using Flan-T5 models: 76M, 250M, and 783M, which are available at https://huggingface.co/google. Additionally, we also tested with larger models such as Llama2 and ERNIE-Bot-4. Llama2 encompasses versions: Llama2 7b, and Llama2 70b. Our approach was compared with state-of-the-art zero-shot baseline models,using F1 score as the evaluation metric for zero-shot testing.The Flan-T5 model was executed on a V100 GPU during the experiments, while the Llama2 models were tested on the Baidu Intelligent Cloud Qianfan platform. This platform is an all-in-one enterprise-level platform for large models, providing advanced generative AI production and application development toolchains. The latest version of the Wenxin large model 4.0 has been officially released, offering users powerful functionality and a convenient development environment. In the implementation process, we fully utilized the advantages of the LangChain large model application development tool. By customizing the LLM language model class, we seamlessly integrated different large-scale language models. By leveraging the LangChain tool, we were able to efficiently develop and integrate various LLM models, thus ensuring the functionality of the system. This customizable approach provides more options and effectively addresses various language processing tasks, providing a reliable and efficient foundation for this research.

\subsection{Inference Results} 
Table 2 presents a comprehensive comparison of various models employed for zero-shot reasoning. When applied to the restaurant dataset, the ERNIE-Bot-4+SAoT approach achieves an F1 score of 75.27, accompanied by an ISA score of 66.29. On the computer dataset, it attains an F1 score of 76.50, and an ISA score of 73.46.

In comparison to the BERT$_{Asp}$+SCAPT model, the ERNIE-Bot-4+SAoT model demonstrates notable advancements. It exhibits a substantial increase of  45.25\% ( = 75.27 - 30.02) in the F1 score  on the Restaurant dataset and an impressive improvement of  50.73\% ( = 76.50 - 25.77) on the Laptop dataset. On average, there is an overall improvement of  47.99\% ( = (45.25 + 50.73) / 2) .

Additionally, we performed sentiment analysis on the Llama2+SAoT (7B) and Llama2+SAoT (70B) model series. we observed a direct correlation between the model parameter size of Llama2 and the enhancement in the F1 score. As the model parameter size increased, the F1 score improved accordingly.

\begin{longtblr}[
  caption = {Zero-shot Performance Evaluation of the Models}, 
]{
  width = \linewidth,
  colspec = {Q[483]Q[121]Q[121]Q[100]Q[102]},
  cell{1}{2} = {c=2}{0.242\linewidth},
  cell{1}{4} = {c=2}{0.202\linewidth},
  cell{3}{1} = {c=5}{0.926\linewidth},
  cell{7}{1} = {c=5}{0.926\linewidth}, 
   cell{22}{1} = {c=5}{0.926\linewidth},
  hline{1,3-4,7-8,12-13,22-23} = {-}{},
  hline{2} = {2-5}{},
}
 & Restaurant &  & Laptop & \\
 &  F1 &  ISA &  F1 & ISA\\
• State-of-the-art baselines &  &  &  & \\
BERT+SPC$^{\dag}$  & 21.76 & 19.48 & 25.34 & 17.71\\
BERT+RGAT$^{\dag}$  & 27.48 & 22.04 & 25.68 & 18.26\\
BERT$_{Asp}$+SCAPT$^{\dag}$  & 30.02 & 25.49 & 25.77 & 13.70\\
• Prompt-based methods &  &  &  & \\
Flan-T5+Prompt(76M) & 44.30 & 32.20 & 41.71 & 26.31\\
Flan-T5+Prompt(250M) & 53.06 & 37.19 & 50.43 & 30.29\\
Flan-T5+Prompt(783M) & 56.01 & 38.43 & 54.14 & 35.86\\
ERNIE-Bot-4+Prompt & 55.56 & 36.96 & 53.53 & 36.66\\
• CoT-based methods &  &  &  & \\ 
Flan-T5+THOR (76M) & 46.73 & 33.51 & 41.89 & 25.55\\
Flan-T5+THOR (250M) & 53.32 & 36.35 & 50.89 & 26.85\\
Flan-T5+THOR (783M) & 55.02 & 38.77 & 52.08 & 32.15\\
 
Flan-T5+SAoT (76M) & 48.61 & 36.81 & 46.34 & 28.19\\
Flan-T5+SAoT (250M) & 57.49 & 37.36 & 54.21 & 34.14\\
Flan-T5+SAoT (783M) & 62.43 & 48.18 & 64.31 & 50.39\\
Llama2+SAoT (7B) & 60.77 & 50.41 & 67.03 & 59.37\\ 
Llama2+SAoT (70B) & 68.52 & 54.51 & 66.59 & 54.98\\
ERNIE-Bot-4+SAoT & \textbf{75.27} & \textbf{66.29} & \textbf{76.50} & \textbf{73.46}\\
 
\TableNote{Scores with  $^{\dag}$ were obtained from \cite{prompt-thor}.}
\end{longtblr}

\begin{figure}[h]
  \centering
  \includegraphics[width=\linewidth]{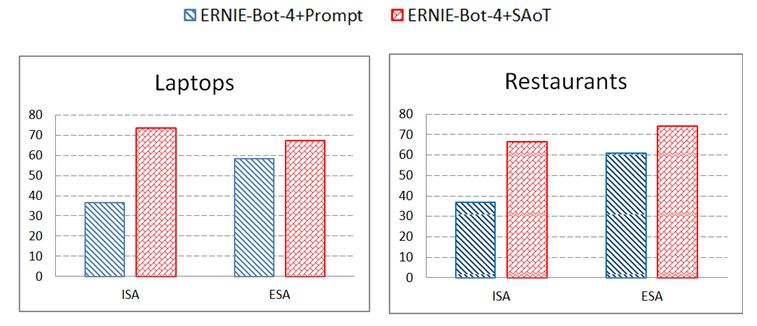}
  \caption{Comparing ERNIE-Bot-4 with Prompt and SAoT for ESA and ISA Instances}
 
\end{figure}
    
\subsection{Enhancing Performance with SAoT Applications} As shown in Figure 2, we compared the performance differences between the prompt method and the SAoT method based on the ERNIE-Bot-4 model. The outcomes of the analysis on the Laptop and restaurant datasets are illustrated in Figure 2. It can be observed that both the prompt and SAoT methods exhibit relatively high performance levels in ESA, although the improvement achieved by SAoT is limited. However, the prompt-based method shows a higher failure rate in ISA, while ERNIE-Bot-4+SAoT has achieved significant improvements in ISA.

\section{Conclusion} 
This paper presents a sentiment analysis approach that leverages the SAoT framework, aiming to achieve the inference process of implicit sentiment analysis reasoning chains. Building upon existing large-scale language models, we design corresponding prompts for the inference steps of SAoT. By leveraging commonsense and the ability to reason through thinking chains, the large-scale language model performs inferential analysis on implicit aspects and opinions in the text, reflecting on the process of implicit sentiment analysis. Ultimately, we are able to infer the polarity of sentiment. On the Laptop dataset, Our SAoT approach demonstrates superior performance in the zero-shot setting of the ERNIE-Bot-4 model. 

\section*{Acknowledgments}
The sponsorship for this research is provided by the Baidu Intelligent Cloud Qianfan Big Model Platform in China, specifically the Wenxin Big Model 4.0. ERNIE-Bot-4, a large language model, has been developed by Baidu, which covers a vast amount of Chinese data and possesses stronger capabilities in dialogue-based question answering, content generation, and more.

\bibliographystyle{IEEEtran}

\end{document}